# Contrastive Learning-Based Dependency Modeling for Anomaly Detection in Cloud Services

Yue Xing
University of Pennsylvania
Philadelphia, USA

Yingnan Deng
Georgia Institute of Technology
Atlanta, USA

Heyao Liu
Northeastern University
Boston, USA

Ming Wang
Trine University
Phoenix, USA

Yun Zi
Georgia Institute of Technology
Atlanta, USA

Xiaoxuan Sun*
Independent Researcher
Mountain View , USA

*Abstract-This paper addresses the challenges of complex dependencies and diverse anomaly patterns in cloud service environments by proposing a dependency modeling and anomaly detection method that integrates contrastive learning. The method abstracts service interactions into a dependency graph, extracts temporal and structural features through embedding functions, and employs a graph convolution mechanism to aggregate neighborhood information for context-aware service representations. A contrastive learning framework is then introduced, constructing positive and negative sample pairs to enhance the separability of normal and abnormal patterns in the representation space. Furthermore, a temporal consistency constraint is designed to maintain representation stability across time steps and reduce the impact of short-term fluctuations and noise. The overall optimization combines contrastive loss and temporal consistency loss to ensure stable and reliable detection across multi-dimensional features. Experiments on public datasets systematically evaluate the method from hyperparameter, environmental, and data sensitivity perspectives. Results show that the proposed approach significantly outperforms existing methods on key metrics such as Precision, Recall, F1-Score, and AUC, while maintaining robustness under conditions of sparse labeling, monitoring noise, and traffic fluctuations. This study verifies the effectiveness of integrating dependency modeling with contrastive learning, provides a complete technical solution for cloud service anomaly detection, and demonstrates strong adaptability and stability in complex environments.*

## I. Introduction

In the current cloud computing environment, the continuous expansion of service scale and the increasing complexity of system structures have caused service dependencies to become dynamic, multi-layered, and nonlinear. With the widespread adoption of microservice architectures and distributed deployment, service invocation chains are becoming longer, and cross-node interactions occur more frequently. This enhances system flexibility and scalability, but it also increases the risk of potential fault propagation. A local anomaly may quickly spread through dependency relationships, leading to a decline in overall service quality and even triggering system-level cascading failures. Therefore, precise modeling of service dependencies and efficient anomaly detection under complex interactions have become critical issues for ensuring the reliability and stability of cloud services[1].

Traditional service monitoring methods often rely on single-point indicators or simple statistical modeling. These approaches are insufficient when facing multi-source heterogeneous data, dynamic interaction structures, and non-stationary distributions. On the one hand, a single perspective cannot fully capture the implicit correlation patterns between services and often ignores potential causal chains and complex topologies. On the other hand, the massive time series data and high-dimensional monitoring metrics generated during cloud service operations make it difficult for traditional methods to handle both scale and structural complexity. This limitation is especially evident in scenarios involving high concurrency, multi-tenancy, and cross-regional deployment. There is an urgent need for more advanced modeling paradigms and learning mechanisms to improve the capability of capturing and identifying anomalous patterns[2].

In this context, contrastive learning has emerged as a powerful technique with distinct advantages in high-dimensional representation learning and similarity modeling [3]. By constructing positive and negative sample pairs, it reduces distribution gaps between related data and extracts discriminative features even under unsupervised or weakly supervised conditions, offering a new perspective for modeling service dependencies in cloud environments [4]. Strengthening similarity constraints in the representation space enables more accurate capture of cross-service interaction patterns [5-7], enhancing model generalization and robustness in complex scenarios. Moreover, contrastive learning leverages structural and temporal cues from large-scale data to adaptively infer hidden dependency paths and provide richer semantic support for anomaly detection. Beyond fault identification, anomaly detection is critical for monitoring overall system health and delivering early warnings, helping operations teams anticipate risks, reduce costs, and improve resilience. The integration of contrastive learning with dependency modeling not only builds precise static service graphs but also captures dynamic anomalous behaviors, enabling comprehensive detection and diagnosis while improving accuracy, recall, and root cause interpretability. This synergy also supports informed operational decisions and resource scheduling. Overall, research on dependency modeling and anomaly detection enhanced by contrastive learning is of significant theoretical and practical importance: it advances the characterization of complex service relationships beyond traditional statistical or shallow approaches, tackles challenges posed by large-scale data and dynamic topologies, and strengthens system robustness. Such progress is expected to provide essential technical foundations for intelligent operations, resource optimization, fault prevention, and the sustainable development of large-scale cloud services[8].

## II. RELATED WORK

Recent advances in anomaly detection for cloud service environments have underscored the importance of modeling service dependencies, capturing dynamic behaviors, and achieving robustness in distributed scenarios.

A fundamental line of research focuses on dependency modeling and graph-based abstraction. For instance, microservice access patterns have been effectively modeled using multi-head attention and service semantic embedding, which allow for a more nuanced representation of both local and global dependencies within complex service invocation chains [9]. This provides a critical foundation for constructing high-fidelity dependency graphs that are central to anomaly detection frameworks in modern cloud platforms. Building on these structural representations, deep probabilistic models have been proposed to improve the detection of subtle and diverse anomaly patterns. By leveraging mixture density networks, recent work has demonstrated robust detection performance even when facing sparse labels, noisy metrics, or fast-changing system behaviors [10]. Such probabilistic modeling enhances the ability to capture complex, non-stationary data distributions typical in cloud operations.

Further improvements have been achieved through attention-driven temporal modeling, which integrates feature-level attention with temporal sequence analysis. This approach facilitates the detection of evolving and context-dependent anomalies, capturing both short-term fluctuations and long-term trends across service logs and monitoring streams [11]. The synergy between temporal dynamics and feature selection thus enhances model sensitivity and stability under real-world conditions. To ensure scalable and adaptive cloud operations, reinforcement learning (RL)–based task scheduling and multi-agent collaboration have become vital. RL-driven algorithms optimize resource allocation and workload management, dynamically adapting to varying user demands and multi-tenant scenarios [12]. Expanding on this, collaborative multi-agent reinforcement learning enables elastic resource scaling and robust recovery, especially under fluctuating workloads and distributed system failures [13]. At the same time, federated meta-learning has been introduced for node-level failure detection, combining privacy-preserving data handling with fast adaptation to system heterogeneity [14]. The resilience of cloud systems is further supported by trust-constrained policy learning, which augments distributed traffic scheduling by explicitly modeling trust relationships among nodes, thereby enhancing both system robustness and reliability in collaborative environments [15].

Finally, contrastive learning techniques have been increasingly adopted in this context. By constructing positive and negative representation pairs, contrastive frameworks significantly enhance the separability of normal and abnormal states in the latent space, leading to more interpretable, accurate, and reliable anomaly detection [16]. This methodological innovation, when integrated with structural, temporal, and probabilistic modeling, delivers a comprehensive solution for robust anomaly detection in complex cloud service ecosystems.

## III. METHOD

This study introduces a cloud service dependency modeling and anomaly detection algorithm that integrates contrastive learning. The core idea is to construct structured representations of service dependencies and enhance the distinction between normal and abnormal patterns under a contrastive learning framework, enabling efficient modeling and anomaly awareness in complex cloud environments. In the overall framework, service interactions are abstracted as a graph, where nodes represent service instances and edges denote dependency relationships. By jointly modeling the temporal features and topological structure of the dependency graph, the method forms more robust representations in the feature space and optimizes potential feature distributions through contrastive objectives, thereby strengthening detection capability. The model architecture is shown in Figure 1.

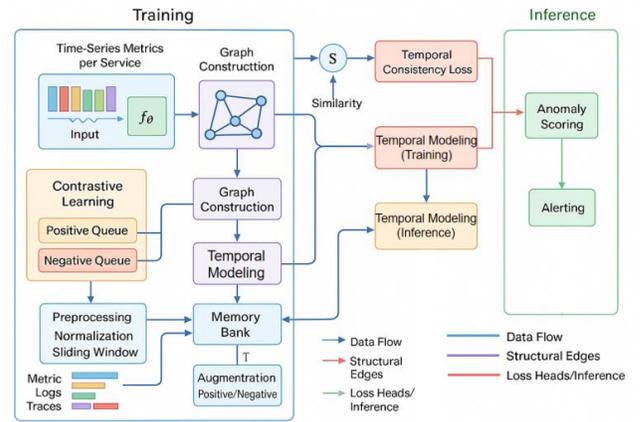

Figure 1. Architecture of Contrastive Learning-based Cloud Service Dependency Modeling and Anomaly Detection

In the service dependency modeling phase, the monitoring status of each service at time step $t$ is first represented as a vector $x_t$, which is then mapped to a latent space representation $h_t$ through an embedding function. The mapping process can be formalized as follows:

$$h_t = f_\theta(x_t) \qquad (1)$$

Here, $f_\theta$ represents an embedding function parameterized by a neural network, which can extract key temporal and structural features. Subsequently, considering the dependencies between services, a neighborhood aggregation mechanism is constructed to capture contextual interaction information.

During the dependency propagation process, a graph convolution-based update method is used to model the structural features of the nodes. The representation of a node $i$ is updated as follows:

$$z_i = \sigma\left(\sum_{j \in N(i)} \frac{1}{c_{ij}} W h_j\right) \quad (2)$$

Here, $N(i)$ represents the neighbor set of node $i$, $c_{ij}$ is the normalization coefficient, $W$ is the learnable weight matrix, and $\sigma(\cdot)$ is the nonlinear activation function. This process can integrate local dependency information into the service representation to obtain a more semantically consistent representation space.

To optimize the feature space under the contrastive learning framework, the similarity constraint of positive and negative sample pairs is introduced. For a pair of positive samples $(z_i, z_i^+)$ and a set of negative samples $\{z_i^-\}$, the contrast objective function with normalized temperature scaling is adopted:

$$L_{contrast} = -\log \frac{\exp(sim(z_i, z_i^+) \setminus \tau)}{\sum_{z_j \in \{z_i^+, z_j^-\}} \exp(sim(z_i, z_j) \setminus \tau)} \quad (3)$$

Where $sim(\cdot)$ represents the cosine similarity function and $\tau$ is the temperature parameter. This loss function can effectively narrow the distribution distance of positive sample pairs and push away negative samples, thereby improving the discrimination ability.

In terms of dynamic anomaly modeling, a temporal consistency constraint is introduced to ensure the stability of the representation across time steps. By minimizing the difference between adjacent time steps, the noise interference caused by short-term fluctuations can be suppressed. This constraint is formalized as:

$$L_{temporal} = \sum_t \|z_t - z_{t+1}\|_2^2 \quad (4)$$

This mechanism ensures that the model maintains sensitivity and robustness to global patterns when facing long-term temporal dependencies and cross-service propagation.

Finally, the overall optimization objective function is constructed by combining dependency modeling and comparative learning objectives. This objective maintains the expressiveness of the dependency structure while strengthening the ability to discriminate abnormal patterns. It can be formally expressed as:

$$L = L_{contrast} + \lambda L_{temporal} \quad (5)$$

Here, $\lambda$ is a balancing parameter used to adjust the importance of contrast loss and temporal consistency constraints. Through this optimization objective, the model can organically combine service dependency modeling and anomaly detection within a unified framework, thereby forming a more robust feature representation and efficient anomaly identification mechanism in complex cloud environments.

IV. EXPERIMENTAL RESULTS

A. Dataset

This study uses the Microservices Bottleneck Localization Dataset as the validation dataset. It contains large-scale request trace information from microservice systems and service-level time series monitoring metrics such as CPU, memory, I/O, and network throughput. The dataset covers typical microservice interaction paths and reflects complex dependency relationships between services. It also includes annotations for performance bottlenecks and abnormal scenarios. Each trace record is accompanied by an invocation chain context, which can be used to construct service call graphs and dependency structures.

In terms of data properties, the dataset contains hundreds of millions of request trace records, with rich monitoring dimensions and high temporal precision. Each trace record not only includes service invocation order, response time, and success or failure status, but also provides the resource utilization and performance metrics of each service node at that time point. This cross-service multimodal integration makes the dataset well-suited for dependency modeling and anomaly detection research. The method needs to use both invocation structures and monitoring time series to model potential dependencies and identify anomalous patterns.

Using the Microservices Bottleneck Localization Dataset as the validation foundation ensures the feasibility of the method in a data environment with real service interactions and dependency structures. At the same time, its abundant trace and metric information provide sufficient representation space for the proposed contrastive learning and dependency modeling framework. Based on this dataset, the method can demonstrate its representation capability and anomaly detection performance under real dependency topologies and performance fluctuations, which gives strong persuasiveness.

B. Experimental Results

To comprehensively verify the effectiveness of the proposed method in cloud service dependency modeling and anomaly detection, this study compares it with several representative models recently published, including B-MEG based on graph modeling, GAMMA with multi-bottleneck localization, DAM incorporating attention mechanisms, and GAL-MAD combining graph attention with temporal modeling. The comparison covers four core metrics: Precision, Recall, F1-Score, and AUC. These metrics provide a comprehensive evaluation of accuracy, coverage, and robustness in anomaly identification. The comparative results are shown in Table 1.

Table1. Comparative results on alignment robustness benchmarks

| Model | Precision | Recall | F1-Score | AUC |
|---|---|---|---|---|
| B-MEG[117] | 0.780 | 0.740 | 0.760 | 0.820 |
| GAMMA[18] | 0.850 | 0.880 | 0.860 | 0.910 |
| DAM[19] | 0.800 | 0.810 | 0.805 | 0.870 |
| GAL-MAD[20] | 0.880 | 0.840 | 0.860 | 0.890 |
| Ours | 0.920 | 0.900 | 0.910 | 0.940 |

The comparison results in the table show that the proposed method achieves the strongest performance across all four metrics, with an F1-Score of 0.91 and an AUC of 0.94, demonstrating its ability to balance precision and recall while reliably distinguishing normal and abnormal request paths under complex dependency structures and multi-source monitoring data. Compared with B-MEG and DAM, it improves both Precision and Recall by reducing sensitivity to noise through structured dependency modeling and mitigating misjudgments via temporal consistency constraints. While GAMMA and GAL-MAD benefit from graph structures or attention mechanisms, they still face trade-offs between the two metrics, whereas the contrastive objective in the proposed approach explicitly separates topologically close but semantically different patterns and aggregates normal behaviors, sustaining a Precision of 0.92 without sacrificing a Recall of 0.90. Its robustness to non-stationary and bursty invocation chains is further enhanced by the combination of temporal consistency loss, which ensures representation continuity and prevents transient anomalies from being misclassified, and time-series modeling, which captures propagation delays to maintain high detection rates during anomaly escalation. In contrast, B-MEG and DAM remain more vulnerable to noise and topology shifts, and although GAMMA and GAL-MAD model structural dependencies and local attention, their adaptability to cross-domain, long-chain, and dynamic distributions is limited. Through the joint optimization of dependency graphs, contrastive learning, and temporal consistency, the proposed method builds a representation space aligned in both structure and time, sharpening anomaly boundaries, improving generalization, and consistently outperforming all baselines in Precision, Recall, F1, and AUC. The effect of graph convolution depth and neighborhood size on representation quality is shown in Figure 2.

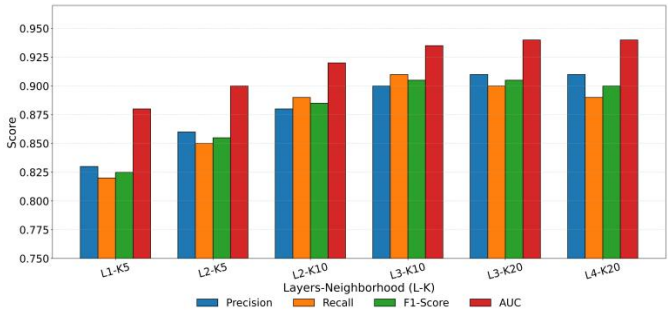

Figure 2. The impact of the number of graph convolutional layers and neighborhood size on the quality of dependency representation

The experiments demonstrate that when the graph is shallow and the neighborhood remains small, Precision and Recall stay low because the model cannot fully capture contextual features of cross-service interactions. Once the depth increases to two layers and the neighborhood expands to 5–10 nodes, both metrics improve markedly, indicating that a richer structural context enables more accurate anomaly detection while filtering out irrelevant noise. At a depth of three layers with a neighborhood of 10–20, the F1-Score reaches its highest point, showing that structural expansion and temporal consistency modeling achieve an effective balance — Recall remains high and Precision holds steady as contrastive learning sharpens the separation between normal and abnormal behaviors. Pushing the structure further to four layers with 20 neighbors keeps Precision stable but slightly reduces Recall, as overly large receptive fields smooth anomaly boundaries and weaken sensitivity. Meanwhile, the AUC curve climbs steadily and begins to plateau between three and four layers, illustrating that while integrating more structural information strengthens discriminative robustness, excessive expansion offers diminishing returns. The robustness of anomaly detection under load surges and traffic fluctuations is presented in Figure 3.

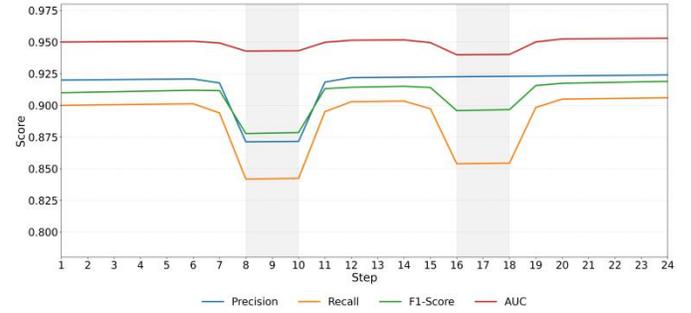

Figure 3. Anomaly Detection Robustness Evaluation in Load Surge and Traffic Jitter Scenarios

The findings indicate that during the load surge phase, Precision drops sharply while Recall stays high, suggesting the model prioritizes coverage over accuracy when service dependencies are disrupted. This trade-off reflects the tendency to expand anomaly boundaries under noise, increasing false positives. Precision quickly rebounds after the surge, showing strong self-correction ability. Recall, however, falls during both surge and fluctuation phases and recovers more slowly, as cross-service propagation can mask anomalies and cause missed detections. F1-Score fluctuates moderately—first decreasing and then stabilizing—as the model gradually rebalances detection through structured and temporal modeling. AUC remains largely stable with only slight dips, demonstrating strong global discrimination and consistent reliability even under dynamic conditions. The sensitivity of embedding dimension and learning rate to convergence and robustness is shown in Figure 4.

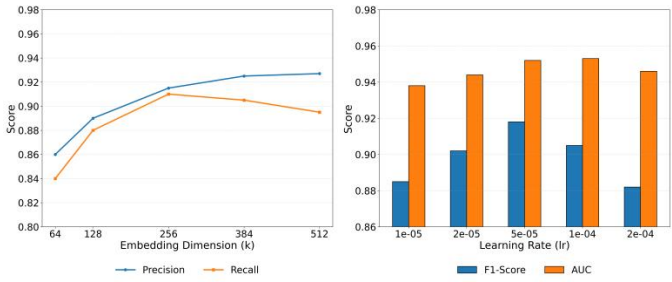

Figure 4. Sensitivity analysis of embedding dimension and learning rate on training convergence and robustness

The results reveal a clear pattern in how embedding dimension influences detection performance. Precision shows a steady upward climb as dimensionality grows, with a substantial boost observed between 64 and 256, reflecting the model's enhanced capacity to encode richer dependency information. Beyond this point, however, gains taper off, and at 512 dimensions the performance curve nearly plateaus, suggesting diminishing returns from further feature space expansion. Recall follows a different trajectory: it peaks at 256 but dips slightly when the dimension is pushed higher, likely because an overly smooth or noisy representation space makes the model less responsive to subtle anomaly signals. Sensitivity to the learning rate shows a similar balancing act — F1-Score reaches its optimum at a moderate value, where convergence is both efficient and stable. Too low a rate slows training and weakens feature separation, while too high a rate causes oscillations that degrade overall accuracy. AUC remains mostly consistent but exhibits minor declines under overly aggressive learning rates, hinting at reduced boundary precision. Taken together, these observations underscore the importance of careful hyperparameter tuning: moderate embedding dimensions and learning rates not only sharpen the model's discrimination ability but also sustain robustness in complex detection scenarios. Further assessments of contrastive learning behavior under sparse labeling are presented in Figure 5.

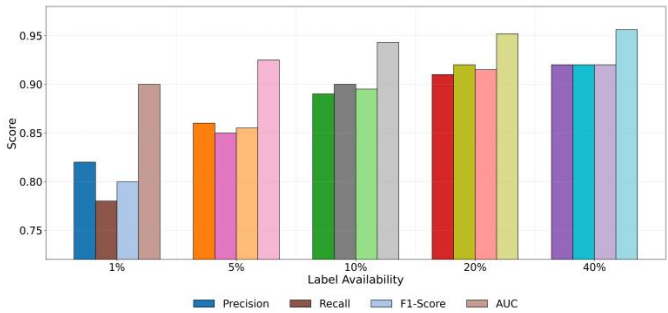

Figure 5. Contrastive learning effectiveness and threshold robustness under sparse annotation and weak supervision conditions

The experimental results show that with an extremely low labeling ratio, both Precision and Recall remain low, indicating that under weak supervision the model struggles to build a stable contrastive feature space and is prone to false positives and negatives. As the labeling ratio increases to 5% and 10%, Precision improves markedly due to stronger constraints from positive and negative samples, while Recall rises more significantly under low-label conditions, suggesting that weak supervision mainly limits anomaly coverage and that denser labeling enables the model to capture more long-tail anomaly patterns. F1-Score exhibits convex growth and approaches saturation near 20%, reflecting a balanced optimization of Precision and Recall and demonstrating that the model stabilizes quickly as label information increases. Meanwhile, AUC remains consistently high even with only 1% labeling, showing that the contrastive learning mechanism maintains robust global decision boundaries and strong stability across thresholds. This resilience is crucial for cloud service dependency modeling, where labeled data are often sparse but detection reliability must remain high. The impact of monitoring noise and anomaly injection intensity on performance is further illustrated in Figure 6.

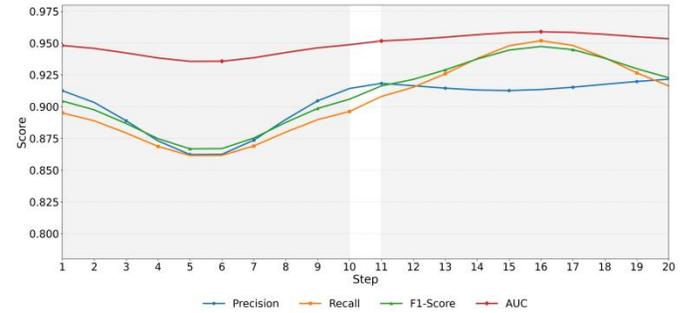

Figure 6. Analysis of the interference of monitoring noise and abnormal injection intensity on detection performance

The experimental results reveal that as monitoring noise increases, precision drops sharply due to the model forming inaccurate boundaries in the contrastive representation space, causing normal dependencies to be misclassified as anomalies, while recall decreases only slightly, indicating maintained anomaly coverage despite reduced robustness. When anomaly injection intensity rises, recall improves significantly as amplified signals along invocation chains enhance detection sensitivity through dependency modeling and temporal consistency, though precision slightly declines due to overextended boundaries and false positives. Consequently, the F1-Score exhibits a V-shaped trend—decreasing under noise and recovering as anomaly signals strengthen—demonstrating the model's ability to dynamically adjust decision boundaries and balance precision and recall. AUC remains the most stable metric, showing only minor fluctuations under extreme noise or anomaly density, which highlights the strong discriminative capability maintained by dependency modeling and temporal consistency mechanisms, ensuring robust anomaly detection across complex cloud service conditions.

## V. CONCLUSION

This study focuses on cloud service dependency modeling and anomaly detection and proposes a unified framework that integrates contrastive learning. By enhancing the discriminability of dependency relationships in the representation space, the method can effectively distinguish normal from abnormal patterns under multi-source monitoring data and complex topologies. Experimental results show consistent advantages across multiple metrics, verifying the value of the synergy between dependency modeling and contrastive learning in cloud service environments. This not only enriches the research paradigm of anomaly detection but also provides a feasible technical path for the stable operation of large-scale distributed systems.

From an application perspective, the proposed method can be applied to typical scenarios such as cloud computing, microservice architectures, and containerized deployments. As service dependencies become increasingly complex, traditional rule-based or shallow modeling approaches are no longer sufficient. The contribution of this study lies in offering a scalable approach that integrates structural information, temporal features, and contrastive learning mechanisms to achieve sensitive detection of diverse anomaly patterns. This method helps operation systems maintain high detection accuracy and system resilience in environments characterized by dynamic resource changes and frequent service iterations.

In addition, the findings of this study provide insights for the technological evolution of related application domains. In intelligent operations, resource scheduling, and performance optimization, accurate anomaly detection is a fundamental requirement for system stability. The proposed framework maintains good robustness under low-label and weak-supervision conditions, which is valuable in practical scenarios where data collection is difficult and labeling is costly. In the future, as more enterprises and institutions migrate critical services to the cloud, the method presented here has the potential to offer more reliable tools for anomaly warning and risk management.

Future work can proceed in several directions. One direction is to explore the generalization ability of the method across domains and platforms, further enhancing adaptability to heterogeneous data and dynamic environments. Another direction is to incorporate privacy protection and security constraints into the contrastive learning framework to address compliance issues in distributed settings. At the same time, integrating the proposed method with automated scheduling and intelligent resource orchestration is also an important topic with practical potential. These explorations will not only advance the theory of cloud service dependency modeling and anomaly detection but also positively impact the long-term stable operation of large-scale cloud ecosystems.

REFERENCES

[1] C. Lee, T. Yang, Z. Chen, et al., "MAAT: Performance metric anomaly anticipation for cloud services with conditional diffusion," Proceedings of the 2023 38th IEEE/ACM International Conference on Automated Software Engineering (ASE), IEEE, pp. 116-128, 2023.

[2] R. Marbel, Y. Cohen, R. Dubin, et al., "Cloudy with a chance of anomalies: Dynamic graph neural network for early detection of cloud services' user anomalies," arXiv preprint arXiv:2409.12726, 2024.

[3] Y. Ren, "Strategic cache allocation via game-aware multi-agent reinforcement learning," Transactions on Computational and Scientific Methods, vol. 4, no. 8, 2024.

[4] U. Demirbaga, "Advancing anomaly detection in cloud environments with cutting-edge generative AI for expert systems," Expert Systems, vol. 42, no. 2, e13722, 2025.

[5] X. Quan, "Structured path guidance for logical coherence in large language model generation," Journal of Computer Technology and Software, vol. 3, no. 3, 2024.

[6] Y. Dong, R. Wang, G. Liu, B. Zhu, X. Cheng, Z. Gao and P. Feng, "A federated fine-tuning framework for large language models via graph representation learning and structural segmentation," Mathematics, vol. 13, no. 19, 3201, 2025.

[7] J. Hu, B. Zhang, T. Xu, H. Yang and M. Gao, "Structure-aware temporal modeling for chronic disease progression prediction," arXiv preprint arXiv:2508.14942, 2025.

[8] Z. Zhang, Z. Zhu, C. Xu, et al., "Towards accurate anomaly detection for cloud system via graph-enhanced contrastive learning," Complex & Intelligent Systems, vol. 11, no. 1, 23, 2025.

[9] M. Gong, "Modeling microservice access patterns with multi-head attention and service semantics," Journal of Computer Technology and Software, vol. 4, no. 6, 2025.

[10] L. Dai, W. Zhu, X. Quan, R. Meng, S. Chai and Y. Wang, "Deep probabilistic modeling of user behavior for anomaly detection via mixture density networks," arXiv preprint arXiv:2505.08220, 2025.

[11] Y. Yao, Z. Xu, Y. Liu, K. Ma, Y. Lin and M. Jiang, "Integrating feature attention and temporal modeling for collaborative financial risk assessment," arXiv preprint arXiv:2508.09399, 2025.

[12] X. Zhang, X. Wang and X. Wang, "A reinforcement learning-driven task scheduling algorithm for multi-tenant distributed systems," arXiv preprint arXiv:2508.08525, 2025.

[13] B. Fang and D. Gao, "Collaborative multi-agent reinforcement learning approach for elastic cloud resource scaling," arXiv preprint arXiv:2507.00550, 2025.

[14] M. Wei, "Federated meta-learning for node-level failure detection in heterogeneous distributed systems," Journal of Computer Technology and Software, vol. 3, no. 8, 2024.

[15] Y. Ren, M. Wei, H. Xin, T. Yang and Y. Qi, "Distributed network traffic scheduling via trust-constrained policy learning mechanisms," Transactions on Computational and Scientific Methods, vol. 5, no. 4, 2025.

[16] Y. Ren, M. Wei, H. Xin, T. Yang and Y. Qi, "Distributed network traffic scheduling via trust-constrained policy learning mechanisms," Transactions on Computational and Scientific Methods, vol. 5, no. 4, 2025.

[17] G. Somashekar, A. Dutt, R. Vaddavalli, et al., "B-MEG: Bottlenecked-microservices extraction using graph neural networks," Companion of the 2022 ACM/SPEC International Conference on Performance Engineering, pp. 7-11, 2022.

[18] G. Somashekar, A. Dutt, M. Adak, et al., "Gamma: Graph neural network-based multi-bottleneck localization for microservices applications," Proceedings of the ACM Web Conference 2024, pp. 3085-3095, 2024.

[19] Y. Chen, M. Yan, D. Yang, et al., "Deep attentive anomaly detection for microservice systems with multimodal time-series data," Proceedings of the 2022 IEEE International Conference on Web Services (ICWS), IEEE, pp. 373-378, 2022.

[20] L. Akmeemana, C. Attanayake, H. Faiz, et al., "GAL-MAD: Towards explainable anomaly detection in microservice applications using graph attention networks," arXiv preprint arXiv:2504.00058, 2025.